# A data-driven approach to predict decision point choice during normal and evacuation wayfinding in multi-story buildings

**Yan Feng**
Transport & Planning Department
Delft University of Technology, Delft, The Netherlands
Email: y.feng@tudelft.nl

**Panchamy Krishnakumari**
Transport & Planning Department
Delft University of Technology, Delft, The Netherlands
Email: P.K.Krishnakumari@tudelft.nl

**ABSTRACT**

Understanding pedestrian route choice behavior in complex buildings is important to ensure pedestrian safety. Previous studies have mostly used traditional data collection methods and discrete choice modeling to understand the influence of different factors on pedestrian route and exit choice, particularly in simple indoor environments. However, research on pedestrian route choice in complex buildings is still limited. This paper presents a data-driven approach for understanding and predicting the pedestrian decision point choice along a chosen route during normal and emergency wayfinding in a multi-story building. For this, we first built an indoor network representation and proposed a data mapping technique to map VR coordinates to the indoor representation. We then used a well-established machine learning algorithm, namely the random forest (RF) model to predict pedestrian's decision point choice during four wayfinding tasks in a multi-story building. Pedestrian behavioral data in a multi-story building was collected by a Virtual Reality experiment. The results show a much higher prediction accuracy of decision points using the RF model (i.e., 93% on average) compared to the logistic regression model. The highest prediction accuracy was 96% for task 3. Additionally, we tested the model performance combining personal characteristics and we found that personal characteristics did not affect decision point choice. This paper demonstrates the potential of applying a machine learning algorithm to study pedestrian route choice behavior in complex indoor buildings.

**INTRODUCTION**

In a building, pedestrians need to find their way to move from one location to another location. During the wayfinding process, their route choice concerns choosing a route when moving from the origin to the destination. When leaving the building, a choice of exit is also required. Understanding pedestrian routes and exit choices during normal and evacuation conditions is crucial to ensure pedestrian safety and evacuation efficiency. Pedestrian route and exit choice are closely linked to the choice of decision points in the building network. Decision points are the locations where pedestrians need to make a choice among multiple available directions or paths along the route (*26*).

Numerous studies have investigated pedestrian route and exit behavior during wayfinding through experiments and modeling. Traditionally, pedestrian route and exit choice data are collected via field experiments (i.e., field observations and lab experiments) and surveys. Applying these methods, a number of studies investigated pedestrian route and exit choice behavior during normal and evacuation situations in different types of buildings (e.g., hospitals, shopping malls, tunnels, and classrooms) (*1–6*). Although these studies provided valuable insights regarding pedestrian route and exit choice behavior, the constraints regarding experimental control, data quality, and ethical and financial restrictions of traditional experimental methods limit the complexity of the experimental scenarios and the type of behavioral data that can be collected. Compared to traditional data collection methods, Virtual Reality (VR) provides possibilities to obtain complete experimental control and collect accurate behavioral data related to route and exit choice behavior in scenarios that are risky or impossible to mimic in the real world (e.g., (*1, 7–14*)). Recently, more studies have attempted to apply VR to understand pedestrian route and exit choices in complex indoor environments (e.g., (*11, 15, 16*)).

With respect to modeling, many studies applied a discrete choice modeling approach to model pedestrian exit choice behavior during evacuation (e.g., (*17–19*)). These decision-making models mostly operate within a framework of rational choice, namely assuming maximum utility in terms of potential outcomes and probabilities of each available option (*20*). Literature shows that the complexity of layout and level changes increases the difficulty of finding one's way in multi-story buildings (*11, 21, 22*). It means in complex buildings with multiple floors and long corridors, pedestrians need to make multiple route choices across vertical and horizontal levels, not just one, which leads to the need of using more decision points. Moreover, due to the complexity of the environmental layout, the difficulty of gathering relevant information also increases. In this case, it is hardly possible for pedestrians to always make the optimal choice. Due to the above-mentioned limitations, most discrete choice modeling studies focused on pedestrian route and exit choices on a single level (i.e., horizontal level) or in simplified environments, modeling pedestrian route and exit choices in multi-story buildings has rarely been studied.

Recently, a few studies have attempted to use machine-learning approaches to study pedestrian behavior. For instance, (*23*) used the random forest model to study pre-evacuation decision-making behavior in a cinema theater. (*24*) combined eye-movement data from real-life environments and employed a random forest model to understand pedestrian wayfinding behavior on a university campus. (*25*) used machine learning methods to investigate pedestrian movement dynamics and exit choice in a multi-exit place. All studies suggested that machine learning methods had better predictive performance compared to traditional modeling approaches. However, to the author's knowledge, no studies have applied machine learning approaches to investigate pedestrian route or decision point choices in multi-story buildings with various task complexities.

To fill this knowledge gap, this study employs a data-driven approach using the Random Forest (RF) model to model and predict pedestrian decision point choice. RF is a machine learning algorithm that is widely used for classification and regression tasks. In this paper, we use the RF model to predict pedestrian decision point choice across various wayfinding task complexities in a multi-story building. To achieve this, we used pedestrian movement trajectory data that was collected via a VR experiment. The VR experiment features four different wayfinding tasks in a multi-story building, providing rich information regarding pedestrian decision point choices. The difficulty and complexity of wayfinding tasks are deliberately increased.

The rest of the paper is structured as follows. Section 2 first presents details of the methodology. Section 3 presents the case study and the RF model results. Finally, we conclude and discuss the future work in Section 4.

## METHODS

In this section, we introduce a data-driven approach to model and predict pedestrian decision point choice in a multi-story building under normal and evacuation conditions. We approach the prediction of decision points as a classification problem.

### 2.1 Indoor building network formulation

The indoor building network is modeled through a hierarchical network consisting of nodes, links, and levels. Using these elements, the representation of an indoor building network can be formulated using a graph. Let G (V, E, L) represent the indoor building network where V represents nodes, E represents links that connect the nodes, and L represents the levels in a building. Here nodes can be understood as decision points, which are the points that pedestrians need to make a choice among multiple available directions or paths along the route (*26*). At decision points, pedestrians need to gather information to determine their next movement during wayfinding. There are three kinds of links, including links that connect nodes on the same level, links that connect two stairways, and links that connect stairways of two levels and access points of another level. The latter two represent the vertical interconnections in the building network. The sequence of links that connect two decision points is generally interpreted as route choices in wayfinding studies. The levels represent the different floors of a building.

### 2.2 Data to decision point mapping

The network of the building and pedestrians' decision point choices are mapped to a directed graph, achieved by a systematic three-step approach. The first step includes precise measurements of the coordinates of the border points within the building network, such as the north, west, south, and east corner of each floor. Additionally, the corresponding coordinates of the above-mentioned nodes in the virtual environment are measured. Based on these coordinates, the second step calculates transformation parameters to convert coordinates between the building network and the virtual environment. The third step includes mapping trajectories onto the network map using the obtained parameters and deriving the decision points usage.

### 2.3 Feature engineering

Feature engineering and selection form the crux of any machine learning project. In our classification task, we systematically engineered new features, selected the most relevant ones, and preprocessed data, including handling categorical variables through techniques like label encoding.

After creating a comprehensive set of features, we implemented correlation analysis and recursive feature elimination as the feature selection technique to identify the most relevant features for our classification task. The objective was to reduce the feature space, prevent overfitting, and improve the interpretability and performance of our model.

To handle categorical features, we used Label Encoding, which involves assigning a unique integer to each category within a feature. This method is straightforward and efficient, although it introduces an arbitrary ordinality among categories. However, this potential issue is less problematic for specific types of models, such as tree-based models, which manage this ordinality effectively. For example, a label encoder for colors could be 0, 1, and 2 to represent 'red', 'green', and 'blue' respectively whereas one hot encoder could be 100, 010, and 001 respectively.

### 2.4 Classification model

There are several data-driven classification methods that can broadly be grouped as linear and non-linear models. In our study, we utilized two primary methods for our classification task: Multiple Logistic Regression and Random Forest. Both methods were carefully chosen, one representing a non-linear approach (Random Forest) and the other a linear one (Multiple Logistic Regression), to provide a comprehensive understanding of our classification problem. Our choice was driven by their efficacy, interpretability, and capability to handle different types of relationships between predictors and the target variable.

*Multiple Logistic Regression*

Logistic Regression is fundamentally a classification method rather than a regression method that models the probability of class membership. It uses Maximum Likelihood Estimation (MLE) to estimate the model's parameters which are then modeled and transformed into probabilities using the logistic function (*27*). The final class membership is obtained by thresholding the probabilities. Despite the simplicity of this method, Multiple Logistic Regression can be a powerful tool for classification when the predictors have a linear relationship with the log odds of the output class.

*Random forest*

Random Forest (RF) is an ensemble learning method that combines multiple decision trees to create a more robust and accurate model (*28*). For each tree, a subset of features is randomly selected at each split, and a majority voting scheme is used to make the final prediction. This technique reduces the risk of overfitting and offers a good balance between bias and variance, making it well-suited for classification tasks. Hyperparameters such as the number of trees in the forest, the maximum depth of the trees, and the number of features considered at each split were fine-tuned using cross-validation to optimize the model's performance.

Moreover, RF provides feature importance metrics, making it a useful tool for understanding the contribution of different features to the prediction. There are different important metrics that can be used. In this work, we use the F-score and the top nodes in the decision trees. The F-score is based on the premise that the more often a feature is used in the different trees of the model, the more important that feature is. In a tree-based model, each decision to split the dataset is made based on a particular feature, and the selection of this feature is generally based on how well it can differentiate or categorize the data. Therefore, the number of times a feature is selected for splitting across all trees in the model could be considered a measure of its importance. The top nodes in a decision tree indicate the features that are important to partition the data most effectively. However, these insights can be limited to a single tree and might not be consistent across the ensemble of trees. Thus, the F-score aggregates information across an ensemble of trees, while the top nodes in a decision tree provide a more granular look at the decision process within a single tree. Both can provide valuable insights into understanding how a model makes its predictions.

## 2.5 Model Evaluation

Assessing the performance of classification models requires suitable evaluation metrics. In our research, we focus primarily on three such metrics: accuracy, balanced accuracy, and the confusion matrix, which provides a comprehensive overview of the model's predictive capability. *Accuracy* is computed as the ratio of correct predictions (both true positives and true negatives) to the total number of predictions. An accuracy score close to 100% signifies a highly predictive model, while a score closer to 0% indicates a poorly performing one. However, accuracy can be misleading in cases of imbalanced classes, where one class significantly outnumbers the others. In such scenarios, a model could predict the majority class for all instances and still achieve high accuracy, despite failing to correctly classify instances from the minority class. Hence, while we utilize accuracy as a primary performance measure, we also use balanced accuracy and the confusion matrix for a more nuanced assessment. *Balanced accuracy* is defined as the average of recall obtained in each class, essentially balancing the metric so that every class contributes equally, regardless of its size (*29*). Balanced accuracy avoids the bias towards the majority class that can occur with regular accuracy and provides a fairer measure of performance on imbalanced datasets. The *confusion matrix* provides an in-depth view of a model's performance (*30*). It is a tabular representation that illustrates the counts of true positive, false positive, true negative, and false negative predictions. This matrix aids in understanding the model's performance on each class, and its tendencies toward false positives (type I errors), or false negatives (type II errors).

**CASE STUDY**

This section presents the data collection (3.1) and results applying the method described in section 2 (3.2) to our dataset.

**3.1 Collection of wayfinding data**

The case study is based on the data collected via a Virtual Reality (VR) experiment that features pedestrian wayfinding behavior in a multi-story building. The VR experiment was conducted from 27th November to 18th December 2019. Ethical approval was obtained from the Human Research Ethics Committee of Delft University of Technology (Reference ID: 944).

The virtual environment features a four-story building that resembles a real-life building with a great level of realism and detail (see **Figure 1**). Each floor (approximately 214 meters long and 12 meters wide) has two major corridors and multiple small corridors that connect the major corridors. There are 8 exits located on the ground floor. During the VR experiment, each participant needed to perform 3 wayfinding tasks under normal conditions and 1 evacuation task. Participants were asked to find their way from room 4.01 to room 4.99 (task 1), from room 4.99 to room 2.01(task 2), from room 2.01 to room 4.64 (task 3), and evacuate to an exit (task 4). These wayfinding tasks were designed deliberately in which the complexity and difficulty of the tasks increased accordingly. After the VR experiment, participants need to fill in a questionnaire.

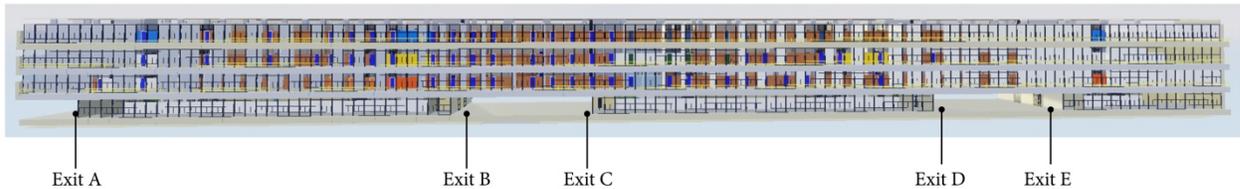

Exit A         Exit B   Exit C        Exit D  Exit E

**Figure 1 An overview of the virtual building**

During the VR experiment, two types of data were collected including subjective data (i.e., movement trajectory) and objective data (i.e., personal characteristics and user experience). Individual movement trajectory was recorded via the VR system with a time interval of 10 milliseconds. It features participants' coordinates (X, Y, Z), head rotations (yaw, roll, pitch), and gaze points (X, Y, Z). For instance, **Figure 2** illustrates the aggregated movement trajectory of all participants. Participants' characteristics (i.e., age, gender, familiarity with the building, highest education level, previous experience with VR, familiarity with any computer gaming) and user experience (i.e., realism, simulation sickness, presence, and usability) are reported via the questionnaire. A previous study by (*16*) evaluated and demonstrated the validity of the behavioral data collected via the VR experiment.

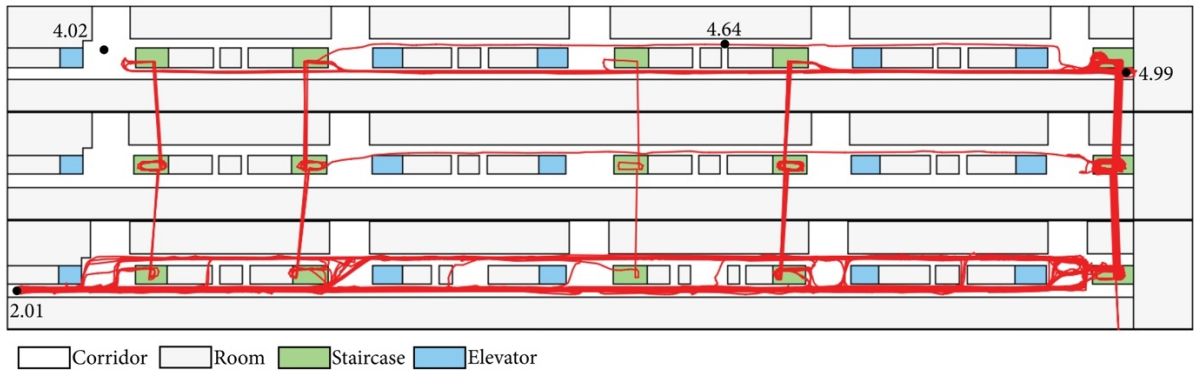

☐ Corridor  ☐ Room  🟩 Staircase  🟦 Elevator

**Figure 2 Aggregated movement trajectory of participants during task 2 (room 4.99 – room 2.01)**

In total, 70 participants finished the VR experiment and considered the data analysis. Among these participants, 34 used the Desktop VR and 36 used the head-mounted-display (HMD) VR to perform the tasks. The participants' age ranged from 17 to 64 years old ($M = 28.27$, $SD = 6.38$). Table 2 presents the personal characteristics of participants. Most participants received a bachelor's degree or higher education level and have a certain level of familiarity with the building. Moreover, 72.85 % of participants had seldom or no experience with VR, and 70.00% of the participants had more experience with computer gaming.

**TABLE 1 Personal characteristics of participants**

| Descriptive information | Category | Number (percentage) |
|---|---|---|
| Gender | Male | 41 (58.57%) |
| | Female | 29 (41.43%) |
| Familiarity with the building | Not at all familiar | 0 (0.00%) |
| | A-little familiar | 6 ( 8.57%) |
| | Moderately familiar | 9 (12.86%) |
| | Quite-a-bit familiar | 16 (22.86%) |
| | Very familiar | 39 (55.71%) |
| Highest education level | High school or equivalent | 5 (13.89%) |
| | Bachelor's degree or equivalent | 16 (22.86%) |
| | Master's degree or equivalent | 40 (57.14%) |
| | Doctoral degree or equivalent | 9 (12.86%) |
| Previous experience with VR | Never | 18 (25.71%) |
| | Seldom | 33 (47.14%) |
| | Sometimes | 15 (21.43%) |
| | Often | 1 ( 1.43%) |
| | Very often | 3 ( 4.29%) |
| Familiarity with any computer gaming | Not at all familiar | 9 (12.86%) |
| | A-little familiar | 12 (17.14%) |
| | Moderately familiar | 13 (18.57%) |
| | Quite-a-bit familiar | 14 (20.00%) |
| | Very familiar | 22 (31.43%) |

## 3.2 Results
### 3.2.1 Indoor network
The indoor network of the multi-story building, comprising nodes (i.e., decision points), links, and levels, was derived following the method outlined in section 2.1 (**Figure 3**). It results in a network consisting of 133 decision points in total. Among these decision points, 10 were exits located on the ground floor and 20 were staircases that connect different floors. Every possible link is connected by two decision points. The numbering of decision points is organized in a way that it starts from the left side of the building and increases the number as moving along the building's layout on that floor. Additionally, to distinguish between the two major corridors, odd numbers are assigned to one corridor, while even numbers are employed for the other corridor. The first digit of the decision point number presents the floor level. To distinguish the staircases, they were numbered using letters A, B, C, D, and E.

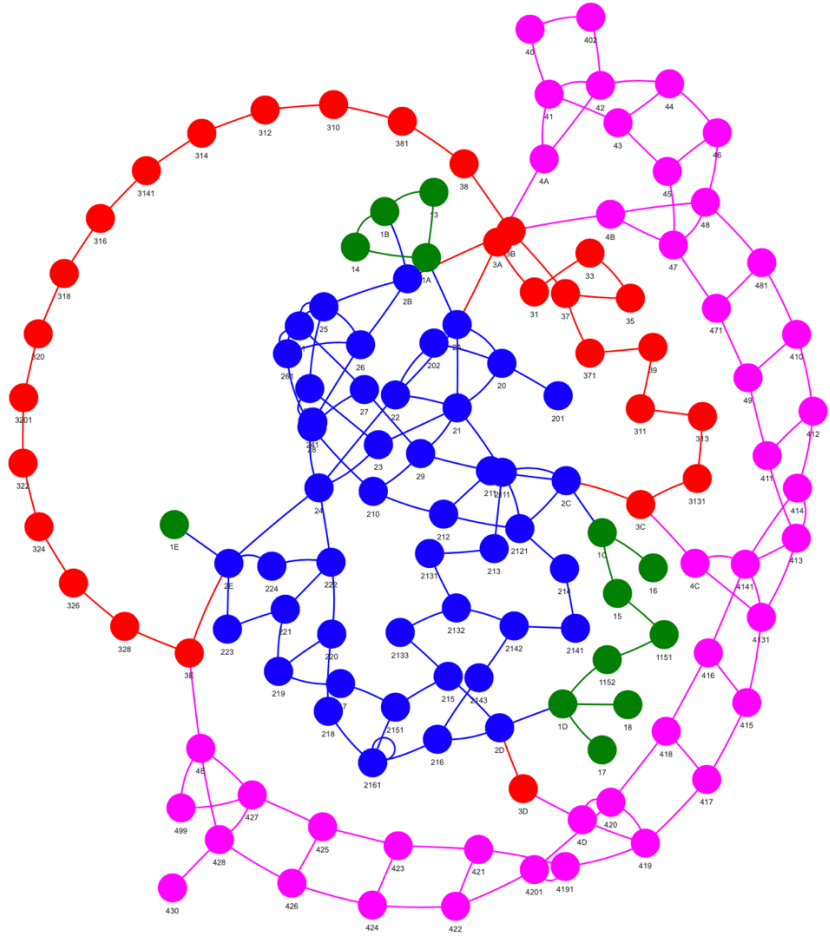

**Figure 3 Illustration of the indoor network, where the pink color represents the fourth floor, the red color represents the third floor, the blue color represents the second floor, and the green represents the ground floor.**

Using the mapping method described in section 2.2, the choice of decision points per task by each participant is mapped to the building's indoor network. **Figure 4** shows the used decision points by one participant during task 1, namely finding one's way from 402 to 499. It shows the mapped nodes and links associated with the person's trajectory.

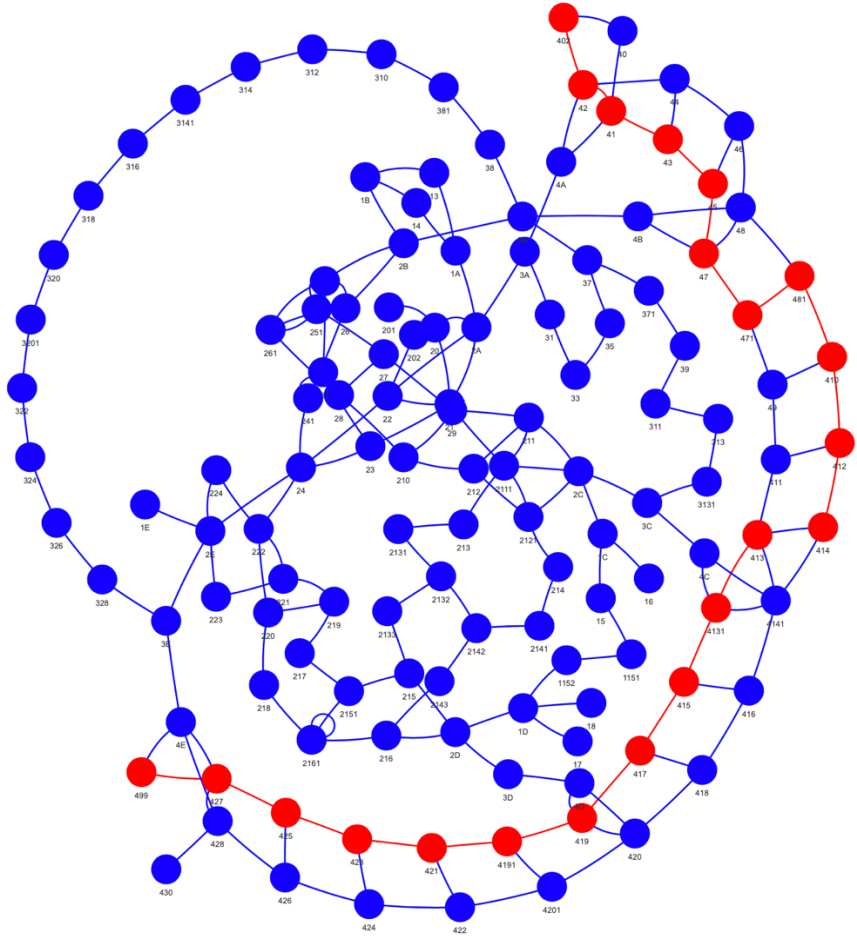

**Figure 4 Example of decision points used during task 1.**

*3.2.2 Preliminary Data Analysis*

The distribution of decision point usage across all tasks of all participants is shown in **Figure 5**. It reveals that besides the starting and ending points of the wayfinding tasks (i.e., 402, 499, 201, 418), the decision points located on the 4th floor were used most frequently. **Figure 6** shows the distribution of the decision points usage per task. For task 1, the decision points located on the same side as room 4.99 were used most frequently. This finding suggests that participants are more inclined to switch to the side where the destination is during wayfinding. For task 2, most participants chose to go to the floor where the destination is located rather than staying on the current floor. Similarly, the majority of the participants chose to ascend to the floor and moved along the corridor where the destination was located during task 3. For the evacuation task (task 4), only exit 15, 16, 17, and 18 were used by participants among 8 available exits, indicating asymmetrical exit choice behavior, as also evidenced in studies (*1, 9*).

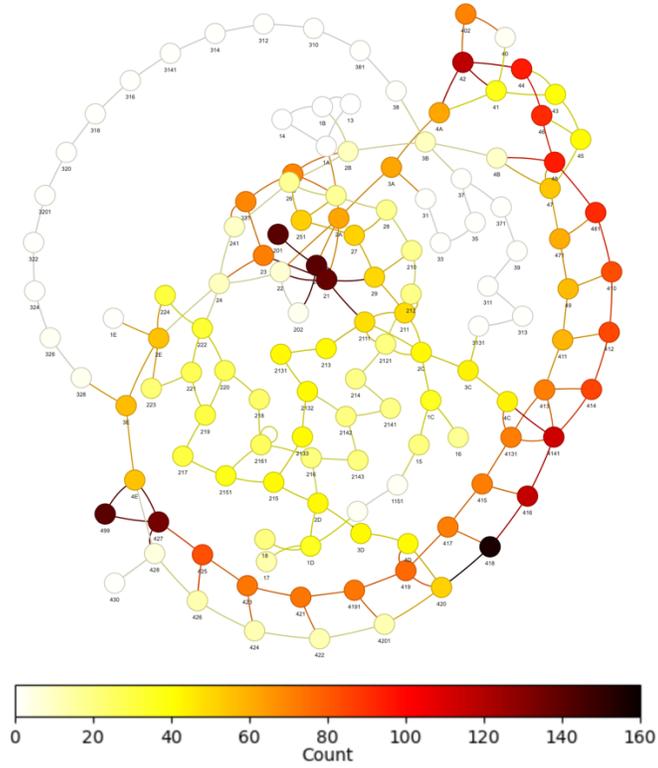

**Figure 5 Density plots showing the distribution of the used decision points by all participants for all tasks.**

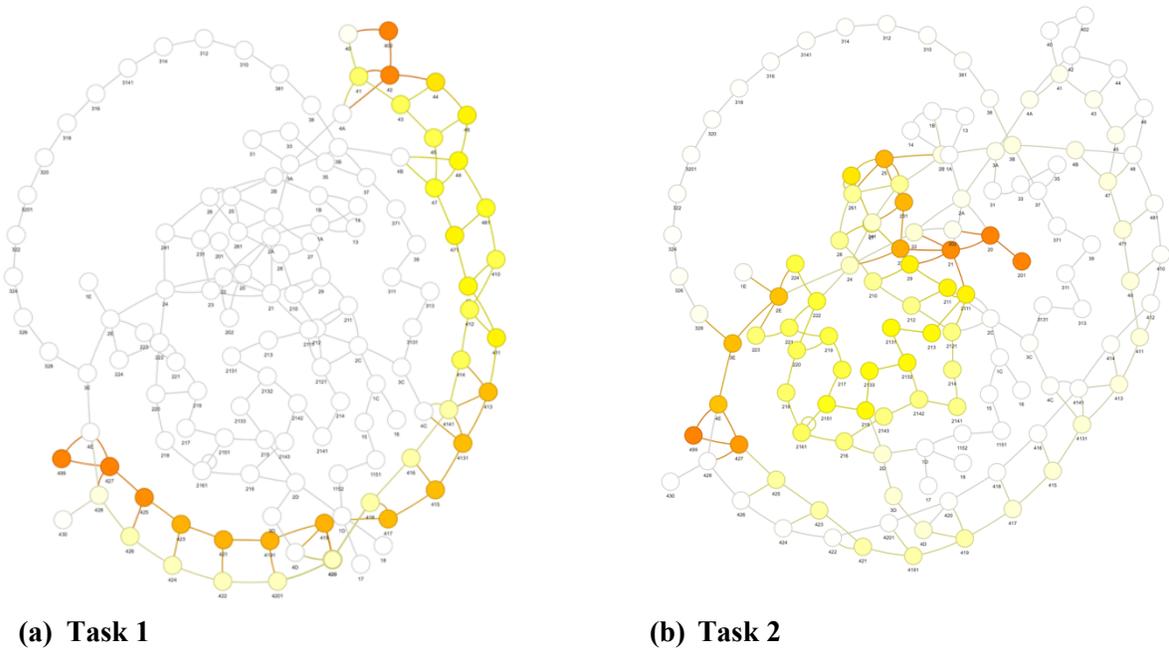

(a) **Task 1**             (b) **Task 2**

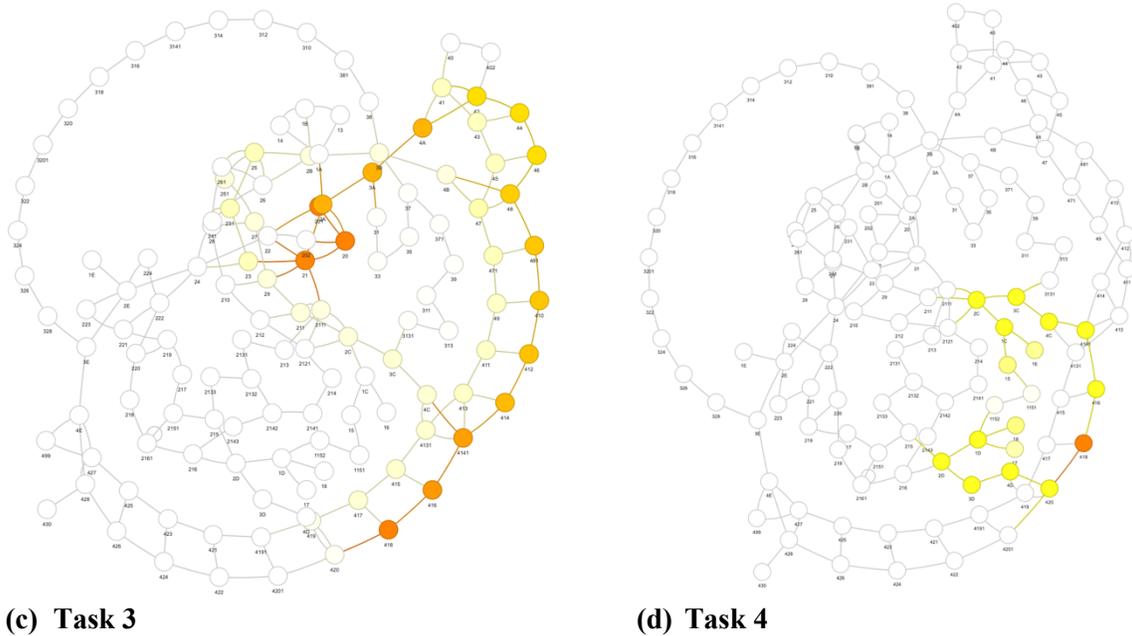

**(c)  Task 3**  **(d)  Task 4**

**Figure 6 Density plots showing the distribution of the used decision points per task.**

The varying number of decision points reflects the difficulties posed by different wayfinding tasks. **Figure 7** presents the distribution of the total number of decision points used per task. On average, participants used 20 decision points for task 1, 26 decision points for task 2, 17 decision points for task 3, and 8 decision points for task 4. The results indicate that, under non-emergency conditions, task 3 demands the fewest decision points to reach the destination, while task 2 demands the highest number of decision points. There are two possible reasons for this: the required travel distance and task complexity. Task 2 required participants to cross both vertical and horizontal levels for the first time to complete the wayfinding task, while task 3 was more straightforward.

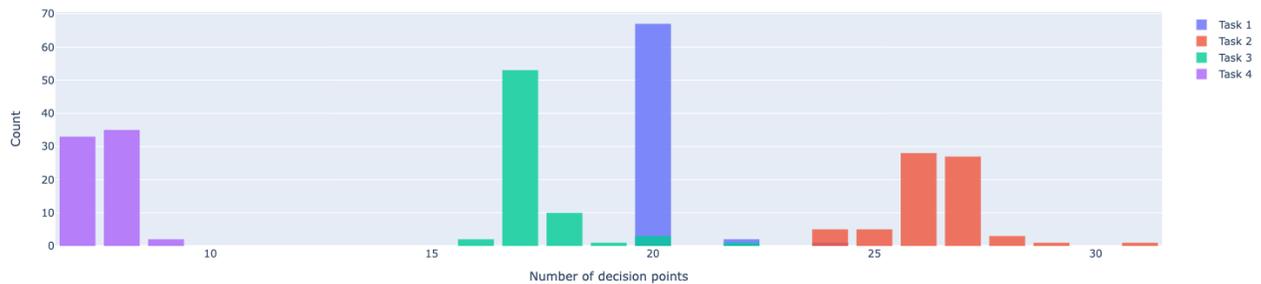

**Figure 7 Distribution of the total number of used decision points per task.**

### 3.2.3 Classification Results

In this section, we present the results of our classification model. First, we compare the performance of the two methods we selected for the classification – MLR and RF. We added the personal characteristics to the RF to investigate if the performance is improved. Then, we discuss in detail the RF results per task. Finally, we end with the sensitivity analysis where we look into the performance with respect to tree depth, number of decision trees, and dimensions of the input.

We transformed the set of mapped consecutive decision points of each person's trajectory of each task into pairwise decision points. The first decision point in the pair is used for the input and the second is the target for the classifier. Thus, the input dimension of the classifier is 2 and is denoted by *[Task number, Previous decision point]*. We used a label encoder to encode the 133 decision points to 133 integers. The inverse transformation can be used to map the labels back to the decision points. The final input and output set contains approximately 4700 data points. We used 80-20 to divide this data into training and testing respectively.

On this data, we applied the MLR and RF and compared their performance. The MLR performed poorly with an accuracy of 5% and balanced accuracy of 2% with a $R^2$ value of 0.07. This is not acceptable and clearly shows that this is not a linear classification problem. On the other hand, the RF model provides an overall prediction accuracy of 93%, indicating a highly precise classification of decision points. The prediction accuracy of each decision point is presented in **Figure 8**.

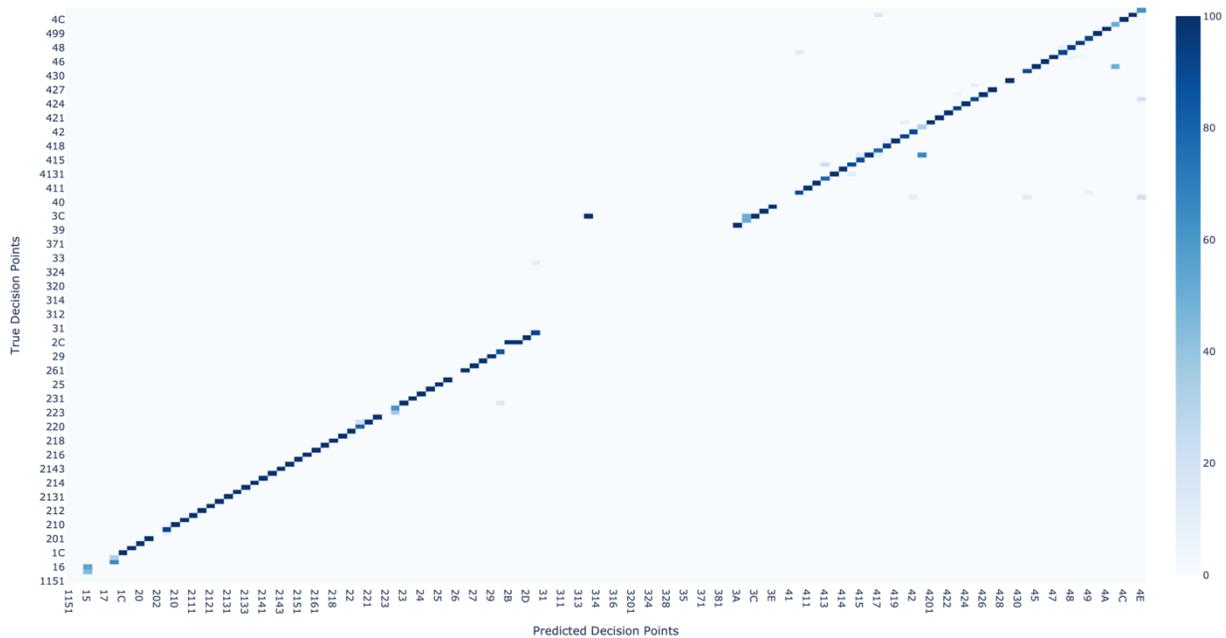

**Figure 8 The results of prediction accuracy for each decision point.**

We expanded the RF classification model to include personal characteristics. We had to use a label encoder for each of the characteristics except the task number, age, and height. We did not normalize it since we are using random forests and they do not have a problem with ordinality. The final input dimension for the classification model is 11 denoted by *[Task number, Previous decision point, Gender, Age, Height, Education, Experience with VR, Experience with Gaming, Familiarity with Building, Experience with Evacuation, Device Type].* Combing the personal characteristics lead to a significantly lower accuracy of 19% and balanced accuracy of 14%. This is in line with the findings in (*19*) which showed that personal characteristics do not contribute to route choice according to discrete choice modeling and the data-driven method supports this. Thus, in the rest of the results, we do not include personal characteristics as input to the model.

We also build a model for each task independently to investigate their individual performance. **Table 2** shows the prediction accuracy and balanced accuracy of each task using the RF model. The results indicate that task 3 has the highest prediction accuracy whereas task 2 has the lowest prediction accuracy. The possible reason for a higher accuracy for task 3 is that it required the least number of decision points for participants to perform the wayfinding task and the variety of decision point choices was relatively lower. **Figure 9** shows the accuracy of each decision point of task 3. Only decision point 251 had a relatively lower accuracy (i.e., 67%) compared to the rest.

**TABLE 2 Prediction accuracy of decision point per task.**

| Task | Accuracy | Balanced Accuracy |
|---|---|---|

| 1 | 95% | 89% |
|---|-----|-----|
| 2 | 93% | 88% |
| 3 | 96% | 89% |
| 4 | 80% | 80% |

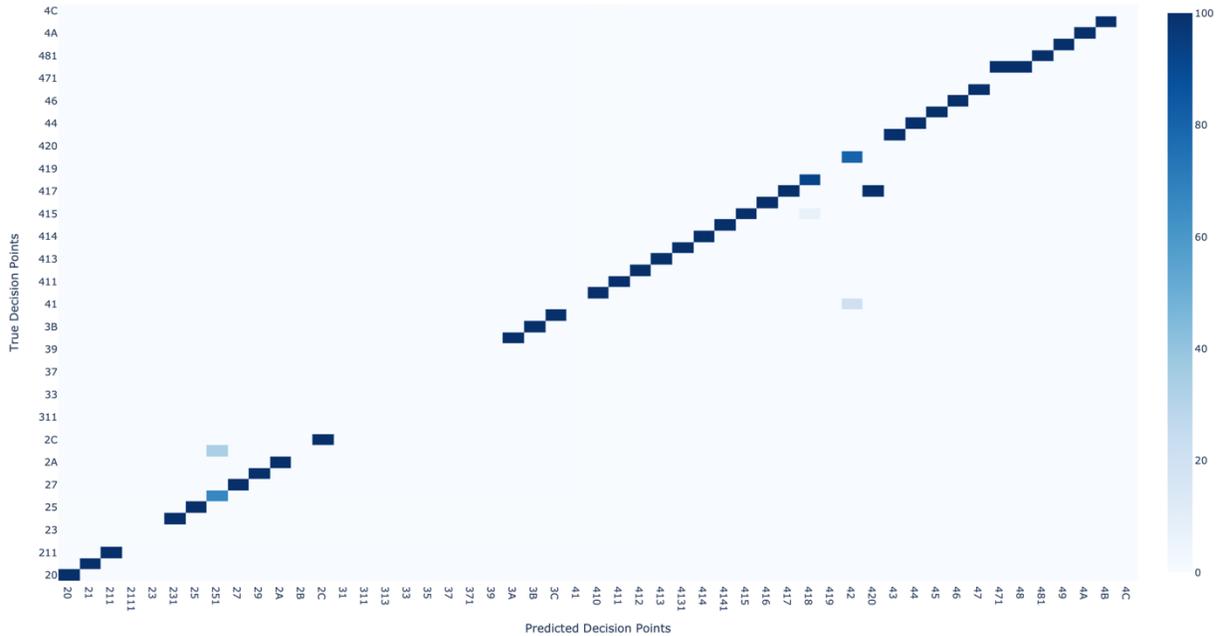

**Figure 9 Prediction accuracy of decision points of task 3.**

The relatively low accuracy of task 4 could be attributed to insufficient input data for predicting exit choice. **Figure 10** shows the accuracy of each decision point of task 4. The decision points with the least accuracy are 416 (i.e., 35%), 16 and 17 (i.e., 0%). In this case, 416 is the first decision point participant needed to choose and the last decision point before participants chose an exit (i.e.,18) is the staircase. This limited information caused difficulties for the RF model to predict whether the participants chose the left-side exit or right-side decision point, which lead to a reduction in overall prediction accuracy for task 4. Moreover, regarding exit 17, we found that the input training data is biased towards 18 than 17 (18 vs 7 input data points).

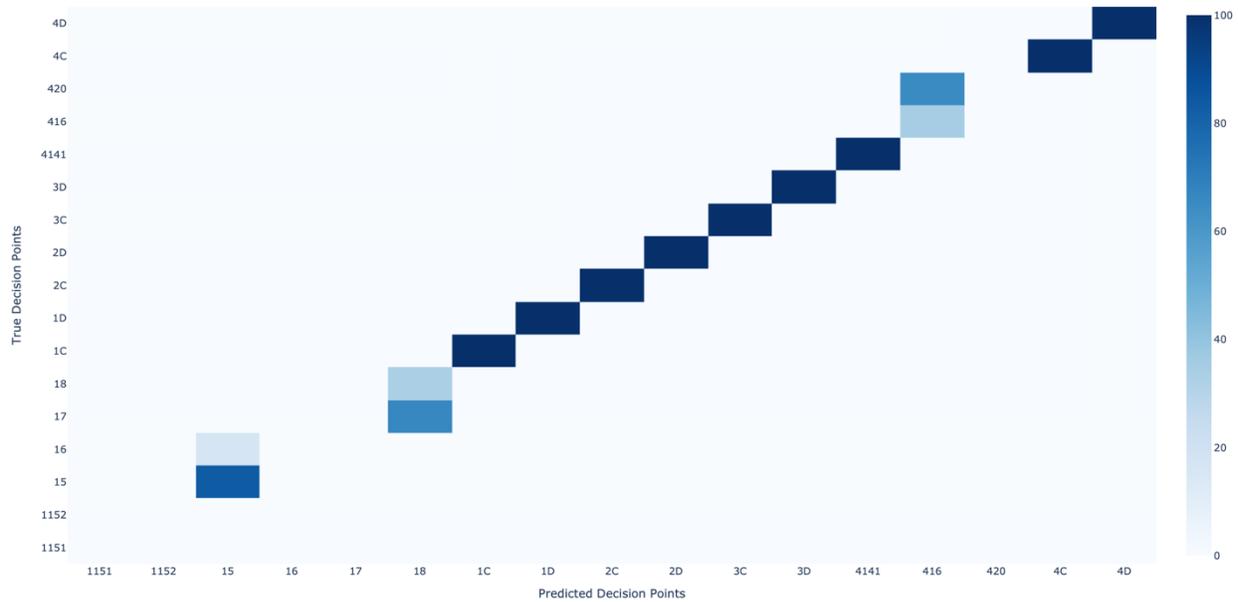

**Figure 10 Prediction accuracy of decision points of task 4.**

Finally, we conducted a sensitivity analysis to tune the parameters of the random forests. We tested the maximum depth parameter from 2 to 40 with a step of 2. We used accuracy as the performance measure. **Figure 11(a)** shows the results, and it can be clearly seen that after a depth of 12, the performance doesn't improve significantly. Thus, we used a maximum depth of 15 for our RF model.

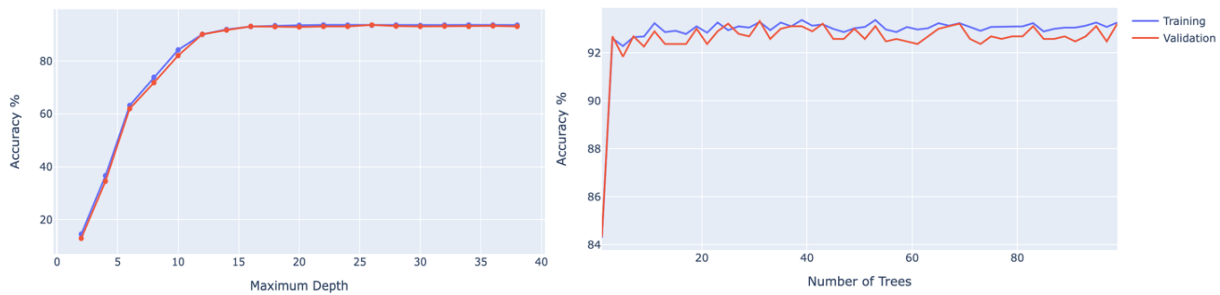

**Figure 11 Sensitivity analysis of random forest for (a) Maximum depth (b) Number of trees**

**Figure 11(b)** shows the results of sensitivity analysis of the number of trees on the accuracy. We tested the performance for the number of trees from 1 to 100 with an increment of 2. The performance is relatively high even for the number of trees = 1, approximately 86%. When the number of trees is 3, the accuracy increases to 93% and it fluctuates slightly as the number of trees increases further. We chose the number of trees as 5 for our RF model.

We also conducted a sensitivity analysis on the input dimension of the RF. We tested the model by providing just the previous decision point and the five preceding decision points, in addition to the task number. The overview of the results is shown in **Table 3** and can be seen that the accuracy doesn't change significantly with increasing dimensions suggesting that the additional information about the person's decision points didn't increase the predictability of the future decision point.

**TABLE 3 Prediction accuracy of decision point for different input dimensions.**

| Input dimension | Accuracy | Balanced Accuracy |
|---|---|---|
| 1 | 93% | 87% |

| 2 | 92% | 81% |
|---|---|---|
| 3 | 93% | 84% |
| 5 | 93% | 84% |

### 3.2.4 Synthesis of the classification results

In this section, we investigate the accuracy of the RF model in detail. First, we try to understand whether the accuracy of the RF model is biased toward personal characteristics. Then, we investigate the feature importance derived from the ensemble trees in RF model and balanced accuracy per decision point.

Regarding gender, the results indicate that the RF model demonstrates a higher prediction accuracy towards female participants (balanced accuracy: 92%) than male participants (balanced accuracy: 85%). We also investigated whether the type of VR device influences prediction accuracy. The results show that the desktop VR group had higher balanced accuracy (90%) compared to the HMD VR group (86%).

With respect to familiarity with the environment, participants who reported moderate familiarity with the building had the highest balanced accuracy (i.e., 96%), and participants who were very familiar with the building had the lowest balanced accuracy (i.e., 86%). **Figure 12** shows the prediction accuracy with respect to familiarity with the building based on per task. It indicates that participants who reported high familiarity had lower balanced accuracy than participants who reported less familiarity during non-emergency tasks. Whereas during the emergency task, people had little familiarity resulting in the lowest prediction accuracy.

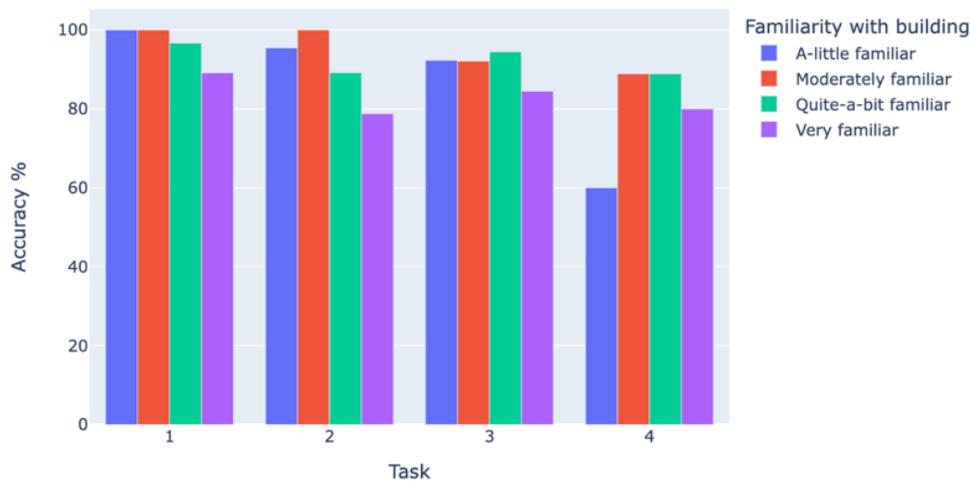

**Figure 12 Prediction accuracy per task with respect to familiarity with the building.**

The feature importance of the RF model is also analyzed to understand the impact of each feature on the RF model's predictive performance. There are two ways to do this as described in the methods section – using the RF F-score and investigating the decision trees. **Figure 13 (a)** shows the F-score for the prediction model with input dimension 2 which includes just the previous decision point and input dimension 6 which includes the preceding 5 decision points as discussed before. **Figure 13 (a), (b)** shows that both models find the 'decision point' as the most important feature for splitting the data than task information. In **Figure 13(b)**, we can see that the importance of the decision point (Decision Point 5) just before the predicted decision point is more important and the importance declines as the preceding decision point is further from the predicted decision point. The 'task' is considered the least important feature for the prediction.

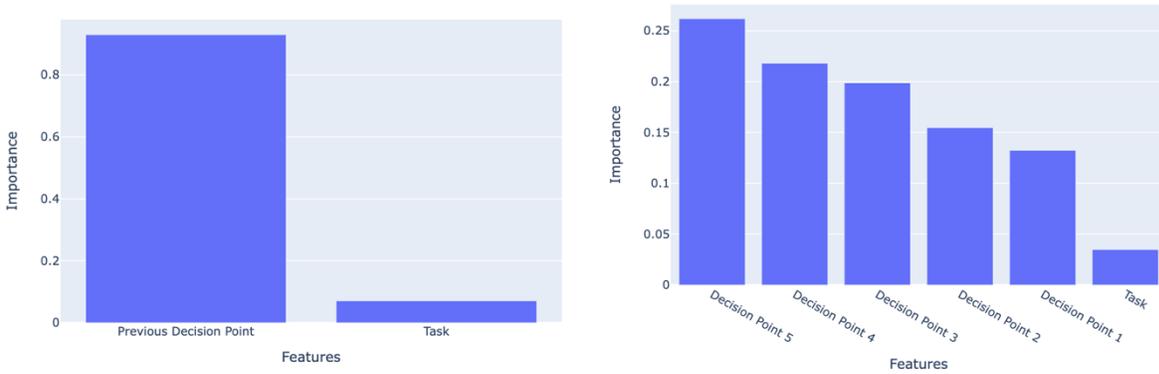

**Figure 13 Feature importance results with (a) one preceding decision point and (b) five decision points as the input to the RF model respectively.**

Moreover, we also analyzed the top nodes of the decision tree. Given that the depth of the tree required for the RF model was 15, which makes the tree unreadable, we only provide some insights here. We investigated the 5 trees in the RF model and consistently found that the top two levels of the trees used 'task' as the feature for splitting the data. This means that even though the 'task' was scored low in feature importance, it was an important initial splitting criterion for the top nodes. Further down the trees, 'decision point' provided more refined splits for the classification. This might lead to 'task' having a lower average feature importance score because this measure averages overall splits in all trees, not just the initial ones. Despite its lower overall importance, 'task' still plays a crucial role in the initial stages of the decision-making process in the trees.

When looking at the predictability of all deicion points across four wayfinding tasks, **Figure 14** shows that the overall predictability of the decision points are quite high, ranging between 80%-100%. However, it is noticeable that decision point 16 and 17 stand out with 0% predict accuracy not only during task 4 but across the entire wayfinding tasks.

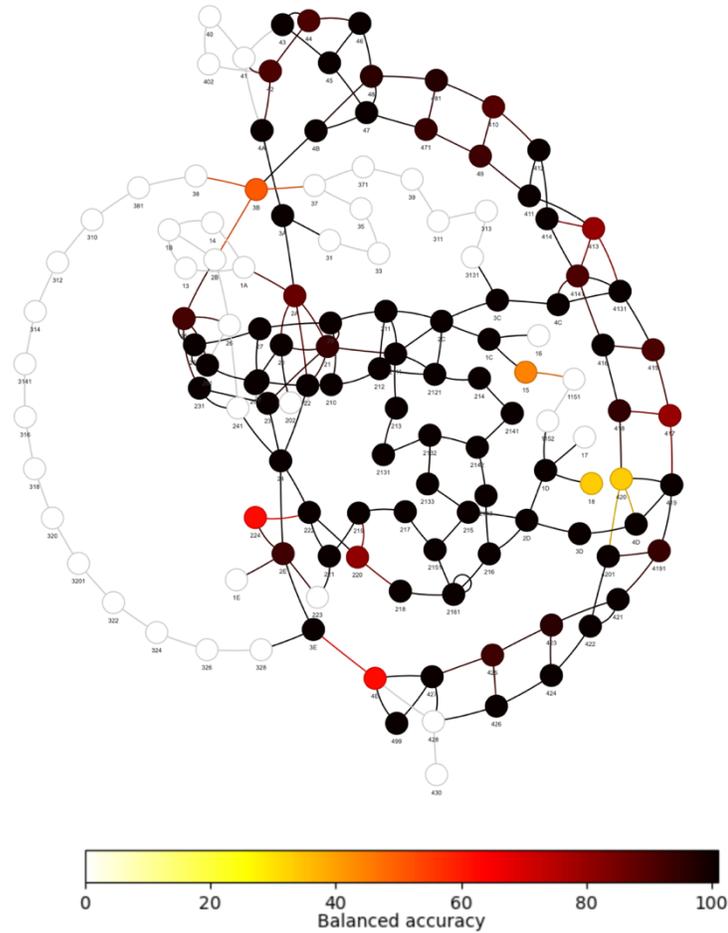

**Figure 14. Overall predictability of decision points.**

**CONCLUSIONS**

In conclusion, this paper investigated pedestrian decision point choice along a route in multi-story buildings using a data-driven approach. To the authors' knowledge, it is the first paper to employ a random forest model to predict pedestrian decision point choices during four wayfinding tasks in a multi-story building. The study presents a thorough approach with respect to data collection, data mapping, feature engineering, and model selection, ultimately employing a Random Forest model for its classification task. The methodology we adopted includes a novel indoor building network formulation to translate the building's layout into a manageable graph, a systemic process of mapping pedestrian decision point choices to a directed graph, and an extensive feature engineering phase to identify the most relevant features for the task. The paper also compares Multiple Logistic Regression and Random Forest models, demonstrating the versatility and robustness of the machine-learning approach.

The results show a high level of prediction accuracy (overall 93%) in predicting the decision point usage in the multi-story indoor network using the RF model, whereas the Multiple Logistic Regression (MLR) model only achieved a 5% accuracy. The RF model, considering tasks individually, achieved the highest accuracy for task 3 (96%) and the lowest for task 4 (80%). Additionally, we found that personal characteristics did not improve the prediction accuracy and were thus excluded in further analysis. Sensitivity analysis showed the RF model performance plateaued at a tree depth of 12 and the number of trees equal to 5. Despite the overall high predictability, decision points 16 and 17 (i.e., two of the used exits) had a 0% prediction accuracy. The analysis also revealed that the RF model has slightly higher prediction accuracy towards female participants and participants using desktop VR. Interestingly, participants reporting moderate familiarity with the building had the highest prediction accuracy, while those very familiar had the lowest. These findings provide significant insights into indoor wayfinding behavior, with practical implications for emergency evacuation planning, indoor navigation system design, and architectural design. This study highlights the potential of using data-driven techniques in contributing to our understanding of pedestrian behavior in complex environments.

There are several limitations in the current study that should be addressed in future research. Firstly, this study did not consider the environmental attributes of decision points, such as their proximity to information points, level of accessibility or visibility. This should be included in future work, which would provide valuable insights into how environmental attributes influence pedestrian decision-point choices. Secondly, the current paper only took participant decision point choices derived from movement trajectories as input data for the decision point prediction. In future studies, richer datasets that feature pedestrian wayfinding behavior (e.g., gaze points) can be considered in order to have a more comprehensive understanding of the decision-making process behind decision point choice. Thirdly, while this paper chose to focus on decision point prediction among other metrics to understand pedestrian route choice in buildings, future research could investigate other metrics considering pedestrian route choice and wayfinding strategies. Furthermore, exploring the transferability of these methods to real-world scenarios will be an interesting implication of this type of study, potentially allowing for more comprehensive and real-time insights into pedestrian movement.

## ACKNOWLEDGMENTS
We would like to acknowledge the utilization of OpenAI's chatGPT-4 in refining the text.

## AUTHOR CONTRIBUTIONS
The authors confirm their contribution to the paper as follows: study conception and design: Y. Feng, P. Krishnakumari; data collection: Y. Feng; analysis and interpretation of results: Y. Feng, P. Krishnakumari; draft manuscript preparation: Y. Feng, P. Krishnakumari. All authors reviewed the results and approved the final version of the manuscript.